\documentclass{article}
\usepackage{iclr2027_conference,times}
\usepackage[T1]{fontenc}

\usepackage{amsmath,amsfonts,bm}

\def\eqref#1{equation~\ref{#1}}
\def\1{\bm{1}}

\DeclareMathAlphabet{\mathsfit}{\encodingdefault}{\sfdefault}{m}{sl}
\SetMathAlphabet{\mathsfit}{bold}{\encodingdefault}{\sfdefault}{bx}{n}

\usepackage{url}
\usepackage{booktabs}
\usepackage{array}
\usepackage{multirow}
\usepackage{amssymb}
\usepackage{algorithm}
\usepackage{algpseudocode}
\usepackage{graphicx}
\usepackage{wrapfig}
\usepackage{framed}
\usepackage{tikz}
\usepackage{hyperref}
\usetikzlibrary{arrows.meta}
\newenvironment{promptbox}{%
  \MakeFramed{\advance\hsize-\width\FrameRestore}%
  \footnotesize\ttfamily\raggedright
  \setlength{\parindent}{0pt}\setlength{\parskip}{0pt}%
}{\endMakeFramed}

\AddToHook{env/table/begin}{\setlength{\belowcaptionskip}{5pt}}
\newcommand{\tablestyle}{\small\renewcommand{\arraystretch}{1.12}\setlength{\tabcolsep}{5pt}}

\title{SelfSearch: Reward-Free Search for\\ Self-Improving Agents}

\author{Jungwoo Yang, Injin Kong, Yohan Jo \thanks{Corresponding author.} \\
Graduate School of Data Science, Seoul National University\\
\texttt{\{jwyang0213,mtkong77,yohan.jo\}@snu.ac.kr}
}

\iclrfinalcopy % Uncomment for the camera-ready version only.

\begin{document}

\maketitle

\lhead{Preprint. }

\begin{abstract}
Advances in the coding capabilities of LLM agents allow them to inspect and modify their own instructions, tools, and execution procedures.
Existing approaches use this ability to search for improved agents through repeated downstream evaluation, which incurs substantial costs and ties the search to the evaluated tasks.
We introduce \textbf{SelfSearch}, a reward-free search procedure in which agents modify themselves using records of previous self-improvement episodes.
These records capture the reasoning, tool actions, and outcomes of earlier modification attempts, providing concrete experience for improving both task solving and self-modification.
Without downstream reward signals during search, SelfSearch improves population-mean success over the initial agent in all six model--benchmark settings, with individual agents gaining up to 11.2 percentage points on Terminal-Bench 2.1.
On SWE-bench Multilingual, an agent improves success by \textbf{5.0} percentage points while reducing execution cost by \textbf{38.5}\% on tasks solved by both the initial and evolved agents.
SelfSearch achieves competitive task success with evaluation-guided search baselines at lower search cost.
With only \textbf{\$4.03} in search cost, it produces a harness that solves \textbf{82.0}\% of Terminal-Bench 2.1 tasks with DeepSeek V4 Flash under the settings of a public nine-harness comparison, matching the top-scoring harness, Codex.
These results suggest that experience gained through self-modification can improve agents' downstream capabilities and efficiency.
\end{abstract}

\section{Introduction}
\label{sec:introduction}

Self-improvement involves recognizing limitations in our capabilities and
developing ways to overcome them. In doing so, we gain experience not only
with the problem at hand, but with how we identify weaknesses, test possible
solutions, and respond to failure. Reflecting on this process can help us
improve how we learn, a central concern of metacognition~\citep{flavell1979metacognition}.
Can agents likewise use the experience of self-improvement to become better
at improving themselves?

The coding capabilities of LLM agents make it possible to investigate this question: agents can inspect and modify the implementations that govern their own behavior.
Agents can learn from their own task-solving experiences and improve their capabilities over time~\citep{zhang2026darwin,wang2026huxleygodel,zhang2026hyperagents}.
These approaches use downstream evaluation to guide the search process,
tying the improvement signal to a task distribution and requiring repeated execution of candidate agents.
Evaluating each candidate on a development set incurs a recurring cost that
grows with the number of tasks evaluated.

Task-solving experience provides evidence about how an agent performs, guiding the search for improvements.
However, a modification produces not only a candidate agent, but also a record of the process that produced it.
This record contains information about how the agent identified limitations, constructed revisions, and examined their effects.
The experience of self-modification can therefore provide valuable evidence for future self-improvement.

We introduce \textbf{SelfSearch}, a reward-free search procedure in which an agent
modifies its own implementation using records of previous self-improvement
episodes. The agent's entire repository is editable, allowing it to revise its
instructions, tools, execution logic, and code organization. A single agent
performs both self-modification and downstream tasks, so changes to its
implementation can affect both capabilities. Each episode produces a successor
agent and a record of the modification process. These records capture how the
agent gathers information, makes changes, and responds to failures. The
successor uses them to guide the next episode, allowing experience and
capabilities developed during self-modification to support further improvement.
For example, difficulty inspecting long records may motivate a reusable
inspection tool, while a failed edit may motivate a procedure that checks the
relevant source before attempting a repair. Later generations can use and
refine both the operations and the procedures developed in earlier episodes.

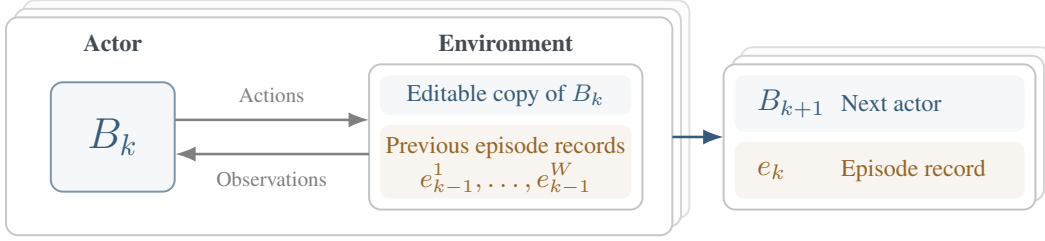
\begin{figure}[t]
\centering
\setlength{\parskip}{0pt}
\setlength{\abovecaptionskip}{4pt}
\resizebox{\linewidth}{!}{% Active manuscript figure. Original preserved in archive/figures/overview_original.tex.
\begin{tikzpicture}[x=1cm,y=1cm,>=Latex,
  every node/.style={font=\small},
  flow/.style={->,line width=0.85pt},
  box/.style={rounded corners=4pt,line width=0.65pt}]
% Keep fixed-size geometry identical in the article preview and ICLR manuscript.
% These definitions are local to the tikzpicture group.
\def\small{\fontsize{9}{10.8}\selectfont}
\def\footnotesize{\fontsize{8}{9.6}\selectfont}
\def\scriptsize{\fontsize{7}{8.4}\selectfont}
\def\normalsize{\fontsize{10}{12}\selectfont}
\def\Large{\fontsize{14.4}{17.28}\selectfont}
\definecolor{actorblue}{HTML}{365B7A}
\definecolor{recordgray}{HTML}{444B53}
\definecolor{experienceamber}{HTML}{946526}
\draw[box,draw=black!12,fill=black!1] (0.20,0.20) rectangle (8.0,2.75);
\draw[box,draw=black!17,fill=white] (0.10,0.10) rectangle (7.9,2.65);
\draw[box,draw=black!22,fill=white] (0,0) rectangle (7.8,2.55);
\node[box,draw=actorblue!65,fill=actorblue!5,minimum width=1.45cm,
  minimum height=1.28cm] at (1.25,1.15) {};
\node[text=actorblue,font=\Large] at (1.25,1.15) {$B_k$};
\node[text=recordgray,font=\footnotesize\rmfamily\bfseries] at (1.25,2.25) {Actor};
\draw[box,draw=black!23,fill=white] (4.25,0.3) rectangle (7.45,2.0);
\fill[experienceamber!7,rounded corners=3pt]
  (4.38,0.43) rectangle (7.32,1.29);
\node[text=recordgray,font=\footnotesize\rmfamily\bfseries]
  at (5.85,2.25) {Environment};
\fill[actorblue!5,rounded corners=3pt]
  (4.38,1.41) rectangle (7.32,1.87);
\node[text=actorblue,font=\footnotesize\rmfamily]
  at (5.85,1.64) {Editable copy of $B_k$};
\node[text=experienceamber,font=\footnotesize\rmfamily]
  at (5.85,1.03) {Previous episode records};
\node[text=experienceamber] at (5.85,0.66)
  {$e_{k-1}^{1},\ldots,e_{k-1}^{W}$};
\draw[flow,black!50] (1.98,1.35) -- (4.25,1.35);
\node[text=black!50,font=\scriptsize\rmfamily] at (3.11,1.64) {Actions};
\draw[flow,black!50] (4.25,0.95) -- (1.98,0.95);
\node[text=black!50,font=\scriptsize\rmfamily] at (3.11,0.66) {Observations};
\draw[flow,actorblue] (7.8,1.15) -- (8.4,1.15);
% Match the environment's vertical bounds and 0.13cm inner padding.
% Output rows have equal height (0.66cm) and a 0.12cm gap.
\draw[box,draw=black!12,fill=black!1] (8.60,0.50) rectangle (12.25,2.20);
\draw[box,draw=black!17,fill=white] (8.50,0.40) rectangle (12.15,2.10);
\draw[box,draw=black!23,fill=white] (8.4,0.30) rectangle (12.05,2.00);
\fill[actorblue!5,rounded corners=3pt] (8.53,1.21) rectangle (11.92,1.87);
\fill[experienceamber!7,rounded corners=3pt] (8.53,0.43) rectangle (11.92,1.09);
\node[anchor=west,text=actorblue,font=\normalsize] at (8.67,1.54) {$B_{k+1}$};
\node[anchor=west,text=actorblue,font=\footnotesize\rmfamily] at (9.65,1.54) {Next actor};
\node[anchor=west,text=experienceamber,font=\normalsize] at (8.67,0.76) {$e_k$};
\node[anchor=west,text=experienceamber,font=\footnotesize\rmfamily] at (9.65,0.76) {Episode record};
\end{tikzpicture}}
\caption{\textbf{SelfSearch.} An agent modifies an editable copy of itself
using previous episode records, which contain interaction trajectories, code
changes, and check results. The successor becomes the next actor, while its
episode record informs subsequent revisions. Stacked boxes represent parallel
lineages. Lineage superscripts are omitted.}
\label{fig:search-selection}
\end{figure}

SelfSearch uses no downstream tasks or evaluation results to guide revisions.
Agents can still use tool outcomes and local checks to inspect and verify their
changes. Downstream evaluation and any candidate selection occur separately
from search.
Our hypothesis is that self-modification exercises capabilities that
are also useful for downstream tasks: inspecting unfamiliar code, diagnosing
failures, implementing changes, and checking their effects. Records of these
activities can therefore provide evidence for improving an agent even before
it encounters downstream tasks.

We evaluate SelfSearch in two model settings on SWE-bench Verified,
SWE-bench Multilingual, and Terminal-Bench 2.1. Each search produces a
population of two agents after ten generations, and we report their mean
performance. The population mean exceeds the base in all six model--benchmark
settings. Individual agents improve success by up to \textbf{11.2} percentage points on
Terminal-Bench 2.1, \textbf{6.7} on SWE-bench Multilingual, and \textbf{5.0} on SWE-bench Verified. On SWE-bench Multilingual, a SelfSearch agent improves success by \textbf{5.0} percentage points while reducing execution cost by \textbf{38.5}\% on tasks solved by both the initial and evolved agents. The discovered
harnesses transfer across model families without further search.

With only
\textbf{\$4.03} in search cost, SelfSearch produces a harness achieving \textbf{82.0}\% on
Terminal-Bench 2.1 using DeepSeek V4 Flash. This matches Codex, the top-scoring harness in a public nine-harness comparison evaluated under the same settings~\citep{apachemaka2026ninearm}.

Our contributions are threefold:
\begin{itemize}
    \item \textbf{Self-improvement as experience.} We introduce SelfSearch,
    which uses episode records to guide agent revisions without
    downstream task evaluations. Each successor becomes the next improver.
    \item \textbf{Capability, efficiency, and transfer.} Across two model
    settings and three benchmarks, SelfSearch improves population-mean task
    success, with execution-cost reductions and harness transfer across model
    families.
    \item \textbf{Understanding how SelfSearch improves agents.} Ablations
    suggest that both previous episode records and an evolving improver
    contribute to population-mean success. Code changes
    and execution traces show how agents use
    self-improvement experience to develop reusable tools and repair failures
    in tool interactions.
\end{itemize}

We also provide an execution framework that enables agent modification within
a controlled boundary, protecting runtime-managed inference settings, resource
limits, and execution records (Appendix~\ref{app:harness-kernel}).

\section{SelfSearch}
\label{sec:selfsearch}

SelfSearch improves agents through a sequence of self-improvement episodes.
In each episode, an agent uses previous episode records to revise its own
implementation. The revised agent performs the next episode, and the new
episode record becomes available to guide subsequent revisions
(Figure~\ref{fig:search-selection}).

\paragraph{Self-improvement episode.}
Let $B_0$ denote the initial agent, implemented as a repository containing its
instructions, tools, and execution logic. The same agent implementation performs
downstream tasks and self-modification, and any part of its repository can be
revised. The underlying model weights remain fixed. In episode $k$, agent $B_k$
receives an editable copy of its repository and read-only episode records $\mathcal E_k$
from previous self-improvement episodes. It examines these records, decides
what to change, and edits and verifies the implementation. The running agent
remains unchanged during the episode. Its edits define the successor $B_{k+1}$,
and the episode produces a record $e_k$:

\begin{equation}
    (B_{k+1},e_k)
    =\operatorname{SelfImprove}(B_k;\mathcal E_k,d).
    \label{eq:selfsearch-transition}
\end{equation}

The search direction $d$ is a qualitative instruction about what kinds of
changes to investigate. The episode record $e_k$ includes the trajectory,
containing the agent's reasoning, tool actions, and their outcomes, together
with the code changes made during the episode. The successor $B_{k+1}$ performs
the next self-improvement episode using previous episode records to guide its
revisions. Changes to its tools and procedures can therefore affect how it
performs subsequent self-modification.

\paragraph{Search across generations.}
Each generation contains one self-improvement episode per lineage. We
maintain two lineages, labeled capability ($c$) and adaptive ($a$). The capability direction asks the
agent to identify limitations in its abilities revealed by inefficient
behavior, failed actions, or difficulty completing an operation, and develop
reusable tools or procedures to address them. The adaptive direction asks
the agent to improve how it revises its approach when actions fail, evidence
contradicts its assumptions, or a better strategy becomes apparent. These
directions guide what agents investigate without scoring the resulting changes.
The two directions encourage agents to investigate different kinds of
limitations.

To give the first generation concrete self-improvement experience to learn
from, we run the initial agent once under each direction to produce the
initial episode records $\mathcal E_0$. We retain these records but discard the edited
implementations, so both lineages begin from the same agent $B_0$. In subsequent
generations, both lineages receive the two episode records from the preceding
generation. Sharing episode records allows each lineage to incorporate
discoveries made under the other direction while maintaining its own
implementation. We retain all generated agents
in the archive $\mathcal C$ for subsequent evaluation.

\begin{algorithm}[t]
\caption{SelfSearch with shared episode records}
\label{alg:selfsearch}
\begin{algorithmic}[1]
\Require Base agent $B_0$, initial episode records $\mathcal E_0$,
search directions $d_{1:W}$, generations $K$
\State $B_0^w\gets B_0$ for each $w\in\{1,\ldots,W\}$
\State $\mathcal C\gets\{B_0\}$
\For{$k=0,\ldots,K-1$}
    \ForAll{$w\in\{1,\ldots,W\}$ \textbf{in parallel}}
        \State $(B_{k+1}^{w},e_k^{w})
        \gets\operatorname{SelfImprove}(B_k^w;\mathcal E_k,d_w)$
    \EndFor
    \State $\mathcal E_{k+1}\gets\{e_k^1,\ldots,e_k^W\}$
    \State $\mathcal C\gets\mathcal C\cup\{B_{k+1}^1,\ldots,B_{k+1}^W\}$
\EndFor
\State \Return candidate archive $\mathcal C$
\end{algorithmic}
\end{algorithm}

\paragraph{Search environment.}
\label{sec:execution-infrastructure}
During search, agents learn from previous self-improvement episodes and
feedback from their current tool interactions and local checks. They receive
no downstream benchmark tasks or evaluation results. We evaluate the resulting
agents only after the search checkpoints have been frozen.

Agents can modify their entire repository, while the model--tool interaction
loop runs in a fixed runtime outside the editable repository. Keeping this loop outside the
editable agent code provides a stable way to execute agents and capture their
trajectories as their implementations evolve. The runtime also controls the
model, reasoning effort, output limits, and execution budgets
(Appendix~\ref{app:harness-kernel}). Agents invoke this loop through a shared
interface and can revise the tools, instructions, and orchestration around it.

\section{Experiments}
\label{sec:results}

\subsection{Experimental Setup}
\label{sec:experimental-setup}

\paragraph{Agents and models.}
We evaluate SelfSearch using two model configurations. The GPT configuration
uses \texttt{gpt-5.6-sol} for self-improvement and \texttt{gpt-5.6-luna} for
downstream execution, both at medium reasoning effort. The DeepSeek
configuration uses \texttt{deepseek-v4-pro-0813} for self-improvement and
\texttt{deepseek-v4-flash-0731} for downstream execution.
Within each configuration, the initial and evolved
agents use the same downstream model and execution limits, differing only
in their agent implementations. Appendix~\ref{app:experimental-details}
provides the full configurations.

\paragraph{Search configuration.}
For each model configuration, we run the two-lineage search described in
Section~\ref{sec:selfsearch} for ten generations. Episodes execute in isolated
containers without network access and support parallel tool calls. We retain
every checkpoint and evaluate the final capability agent $B^{c}$ and
adaptive agent $B^{a}$, reporting their individual results and population mean. We evaluate the final agent from each lineage rather than selecting
checkpoints based on downstream performance.

\paragraph{Downstream evaluation.}
We compare $B_0$, $B^{c}$, and $B^{a}$ on 120 tasks from
SWE-bench Verified~\citep{jimenez2024swebench,openai2024verified}, 60 tasks sampled from
SWE-bench Multilingual covering eight programming languages, and all 89 tasks
in Terminal-Bench 2.1~\citep{merrill2026terminalbench}.
Appendix~\ref{app:experimental-details} provides further evaluation details.

\paragraph{Cost metrics.}
We report average execution cost per task as total model execution cost
divided by the number of attempted tasks, including unsuccessful attempts. Population
success is the mean of the two lineage success rates. We also compare each evolved
agent with the initial agent on tasks solved by both.
Appendix~\ref{app:pricing} defines the cost metrics and pricing, and
Appendix~\ref{app:complete-results} reports token usage and execution costs.

\paragraph{Baselines.}
Our initial agent $B_0$ is derived from DGM's released
implementation~\citep{zhang2026darwin}, with modifications to the policy routing
and execution loop. We use $B_0$ as the baseline for evaluating the evolved
agents. Appendix~\ref{app:harness-kernel} describes the execution infrastructure.
We also compare with two evaluation-guided search baselines: linear search,
which revises the latest candidate, and archive search, which chooses parents
from previously generated candidates. We additionally evaluate two ablations, described in
Section~\ref{sec:results-ablation}.

\subsection{Downstream Performance and Efficiency}
\label{sec:results-performance}

Table~\ref{tab:downstream} compares the initial agent with the capability and
adaptive agents after ten generations. Population-mean success improves in
all six model--benchmark settings. On Terminal-Bench 2.1, the capability
agent increases success from 43.8\% to 55.1\% with GPT and from 65.2\% to
73.0\% with DeepSeek. On SWE-bench Multilingual, the largest gains are
6.7 percentage points with GPT and 5.0 with DeepSeek. Both DeepSeek lineages
improve SWE-bench Verified success from 81.7\% to 86.7\%. The stronger lineage
varies across benchmarks and model settings.

\begin{table}[t]
\caption{Downstream success (\%) and average execution cost per task (USD) for the initial
agent and the capability ($B^{c}$) and adaptive ($B^{a}$) agents after
ten generations. SWE-bench Verified uses the 120-task evaluation set.
Costs exclude search. Bold indicates the best value within
each model and metric, including ties.}
\label{tab:downstream}
\centering
\tablestyle
\renewcommand{\arraystretch}{1.02}
\setlength{\tabcolsep}{5pt}
\begin{tabular}{@{}p{115pt}@{\hspace{6pt}}rrr@{\hspace{16pt}}rrr@{}}
\toprule
 & \multicolumn{3}{c}{\textbf{GPT-5.6 family}} & \multicolumn{3}{c}{\textbf{DeepSeek-V4 family}} \\
\cmidrule(lr){2-4}\cmidrule(lr){5-7}
 & Initial $B_0$ & $B^{c}$ & $B^{a}$
 & Initial $B_0$ & $B^{c}$ & $B^{a}$ \\
\midrule
\csname @@input\endcsname tables/main_combined_rows.tex
\end{tabular}
\end{table}

\par
\ifdim\dimexpr\pagegoal-\pagetotal\relax<330pt
  \newpage
\fi
\begin{wrapfigure}[20]{r}{0.40\textwidth}
\setlength{\abovecaptionskip}{2pt}
\centering
\includegraphics[width=\linewidth]{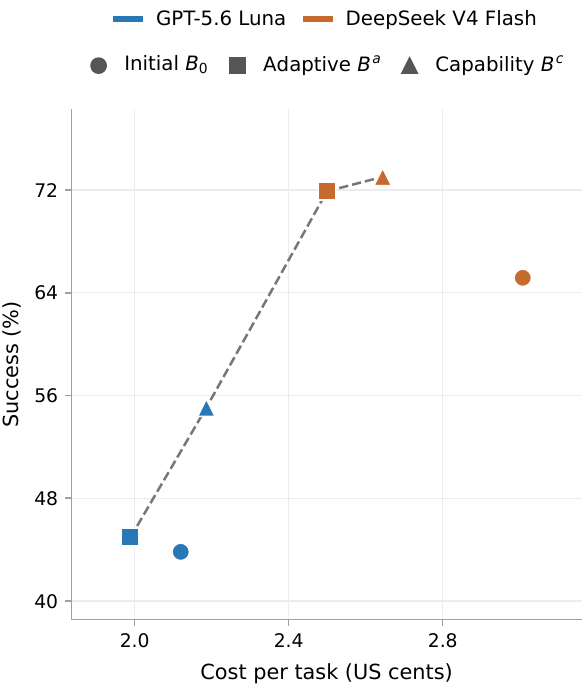}
\caption{Accuracy--cost frontier on Terminal-Bench 2.1. Dashed lines connect
agents on the accuracy--cost frontier.}
\label{fig:terminal-cost-frontier}
\vspace{-8pt}
\end{wrapfigure}
\paragraph{Execution efficiency.}
Figure~\ref{fig:terminal-cost-frontier} compares success rate with average
execution cost per task on Terminal-Bench 2.1. Both DeepSeek lineages improve
success while reducing average cost by 12.1\% and 16.9\% for the capability
and adaptive agents, respectively. The GPT capability agent improves success
from 43.8\% to 55.1\% with a 3.1\% increase in average cost. The adaptive
agent improves success to 44.9\% while reducing average cost by 6.2\%.

We separately compare execution cost on tasks solved by both the initial and
evolved agents. On SWE-bench Multilingual, the DeepSeek adaptive agent
improves overall success by 5.0 percentage points and reduces cost on shared
successes by 38.5\%. On SWE-bench Verified, both lineages reduce cost on
shared successes, by 19.7--36.1\% with DeepSeek and 8.7--13.6\% with GPT.
Appendix~\ref{app:complete-results} provides the corresponding SWE-bench
frontiers, token usage, and detailed cost comparisons.

\subsection{Comparison with Evaluation-Guided Search}
\label{sec:evaluation-guided-comparison}
Both evaluation-guided baselines use a ten-task SWE-bench development
set to guide ten revisions. For each baseline, we select the candidate with
the highest development score, breaking ties in favor of the latest revision.
For this comparison, we exclude the ten development tasks from our 120-task
SWE-bench Verified evaluation set and evaluate all methods on the remaining 110 tasks
alongside SWE-bench Multilingual. SelfSearch retains the
final agent from each lineage without development-score selection.

Table~\ref{tab:evaluation-guided} compares SelfSearch with linear and archive
search on tasks outside the baselines' development set. SelfSearch achieves
competitive success rates without downstream evaluation guiding revisions.
Both DeepSeek lineages match or exceed the baselines on SWE-bench Verified, and the
adaptive lineage matches the strongest baseline on SWE-bench Multilingual. With GPT,
the capability lineage matches or exceeds both baselines on both benchmarks,
while the adaptive lineage shows mixed results.

\begin{table}[t]
\caption{Comparison with evaluation-guided search. Success (\%) and search
cost (USD) on SWE-bench Verified and SWE-bench Multilingual.
The SWE-bench Verified comparison excludes the ten development tasks, leaving
110 tasks. Bold marks the best success rate in each setting.}
\label{tab:evaluation-guided}
\centering\tablestyle
\setlength{\tabcolsep}{3pt}
\renewcommand{\arraystretch}{1.18}
\begin{tabular}{@{}lrrr@{\hspace{10pt}}rrr@{}}
\toprule
 & \multicolumn{3}{c}{\textbf{GPT-5.6 family}} & \multicolumn{3}{c}{\textbf{DeepSeek-V4 family}} \\
\cmidrule(lr){2-4}\cmidrule(lr){5-7}
Method & \footnotesize\shortstack{SWE-bench\\Verified} & \footnotesize\shortstack{SWE-bench\\Multilingual} & \footnotesize Search cost
 & \footnotesize\shortstack{SWE-bench\\Verified} & \footnotesize\shortstack{SWE-bench\\Multilingual} & \footnotesize Search cost \\
\midrule
Initial $B_0$ & 76.4 & 53.3 & --- & 81.8 & 68.3 & --- \\
Linear search & 77.3 & \textbf{60.0} & 12.35 & 81.8 & \textbf{73.3} & 8.59 \\
Archive search & 76.4 & 56.7 & 7.53 & 86.4 & 70.0 & 7.90 \\
\midrule
SelfSearch $B^c$ & \textbf{78.2} & \textbf{60.0} & \multirow{2}{*}{6.52}
 & \textbf{87.3} & 66.7 & \multirow{2}{*}{4.03} \\
SelfSearch $B^a$ & 75.5 & 58.3 & & \textbf{87.3} & \textbf{73.3} & \\
\bottomrule
\end{tabular}
\end{table}

\paragraph{Search cost.}
Generating both SelfSearch lineages costs 13.4--47.2\% less than the baselines
with GPT and 49.0--53.2\% less with DeepSeek. Baseline search costs include editing and development-set evaluation, while SelfSearch search costs cover self-improvement episodes.
Appendix~\ref{app:evaluation-guided} details the comparison protocol.

\subsection{Cross-Model Transfer}
\label{sec:cross-model-transfer}
We evaluate whether the evolved agent implementations remain useful when
executed by a different model, without further search or source-code changes.
Table~\ref{tab:cross-model-transfer} reports transfer in both directions on
SWE-bench Verified. With DeepSeek V4 Flash, the capability and
adaptive agents found using GPT-5.6 Sol achieve 83.3\% and 85.0\% success, compared with 81.7\% for
the initial agent. With GPT-5.6 Luna, the agents found using DeepSeek V4 Pro achieve
79.2\% and 76.7\%, compared with 75.0\% initially. Both lineages improve over
the initial agent in both transfer directions, indicating that the discovered
changes remain useful beyond the model configuration used during search.

\begin{table}[!t]
\caption{Cross-model transfer on SWE-bench Verified (success \%). Rows specify execution
models and column groups specify search models. Superscripts $c$ and $a$
denote capability and adaptive lineages.
Bold marks the best rate within each search model per row, including ties.}
\label{tab:cross-model-transfer}
\centering
\tablestyle
\renewcommand{\arraystretch}{1.02}
\begin{tabular}{@{}l*{5}{>{\centering\arraybackslash}p{43pt}}@{}}
\toprule
 & & \multicolumn{2}{c}{\textbf{GPT-5.6 Sol}} & \multicolumn{2}{c}{\textbf{DeepSeek V4 Pro}} \\
\cmidrule(lr){3-4}\cmidrule(lr){5-6}
Model & Initial $B_0$ & $B^{c}$ & $B^{a}$ & $B^{c}$ & $B^{a}$ \\
\midrule
DeepSeek V4 Flash & 81.7 & 83.3 & \textbf{85.0} & \textbf{86.7} & \textbf{86.7} \\
GPT-5.6 Luna & 75.0 & \textbf{77.5} & 75.0 & \textbf{79.2} & 76.7 \\
\bottomrule
\end{tabular}
\end{table}

\subsection{Comparison with Other Harnesses}
\label{sec:codex-comparison}
We compare the capability agent $B^c$ found using DeepSeek V4 Pro with a public 
evaluation of nine harnesses on Terminal-Bench 2.1 using DeepSeek V4
Flash~\citep{apachemaka2026ninearm}. For this comparison, we align the inference
and execution settings with that evaluation, using \texttt{xhigh}
reasoning and expanded execution limits instead of the settings in
Table~\ref{tab:downstream}. The SelfSearch agent solves 73 of 89 tasks
(\textbf{82.0}\%), tying Codex, the highest-scoring harness in that
comparison. The search cost is \textbf{\$4.03}.
Appendix~\ref{app:published-harness-comparison} provides the evaluation details.

\section{Analysis}
\label{sec:analysis}

We examine the roles of previous episode records and an evolving improver
through ablations, trace how changes develop across generations and pass
between lineages, and inspect the use of evolved tools on downstream tasks.
Here, GPT and DeepSeek refer to
agents found using GPT-5.6 Sol and DeepSeek V4 Pro, respectively.

\subsection{Episode Records and the Evolving Improver}
\label{sec:results-ablation}

\paragraph{Ablation setup.}
We evaluate two ablations of SelfSearch: removing access to previous episode
records and keeping the improver fixed at the initial agent. In the
\textbf{no-record} variant, each revised agent performs the next self-improvement
episode without access to previous episode records. Its implementation carries
forward, so changes to its tools and instructions can accumulate across
generations. In the \textbf{fixed-improver} variant, the initial agent $B_0$
performs every self-improvement episode. It reads previous episode records
and edits the latest implementation in each lineage. The edited implementations
carry forward, but $B_0$ remains unchanged and performs all subsequent edits.
Both variants use the same search configuration as full SelfSearch. We
evaluate the final agent from each lineage on SWE-bench Verified.

\paragraph{Results.}
Full SelfSearch achieves the highest population-mean success in both model
settings (Table~\ref{tab:mechanism-ablation}). Removing episode records reduces
mean success by 2.1 percentage points with GPT and 2.9 with DeepSeek. Keeping
the improver fixed reduces it by 1.7 and 2.9 points, respectively.
With GPT, the fixed-improver
variant performs slightly better in the adaptive lineage.

\begin{table}[!t]
\caption{Mechanism ablations on SWE-bench Verified (success \%). Columns $c$ and $a$
denote capability and adaptive lineages. Mean averages their rates.
Search conditions use generation-10 agents, while Initial reports $B_0$.
Compare within model settings.}
\label{tab:mechanism-ablation}
\centering
\tablestyle
\renewcommand{\arraystretch}{1.02}
\begin{tabular}{@{}lrrr@{\hspace{12pt}}rrr@{}}
\toprule
 & \multicolumn{3}{c}{\textbf{GPT-5.6 family}} & \multicolumn{3}{c}{\textbf{DeepSeek-V4 family}} \\
\cmidrule(lr){2-4}\cmidrule(lr){5-7}
Method & $c$ & $a$ & Mean & $c$ & $a$ & Mean \\
\midrule
Initial agent & 75.0 & 75.0 & 75.0 & 81.7 & 81.7 & 81.7 \\
w/o episode records & 75.0 & 73.3 & 74.2 & 82.5 & 85.0 & 83.8 \\
w/ fixed improver & 73.3 & \textbf{75.8} & 74.6 & 82.5 & 85.0 & 83.8 \\
\addlinespace[3pt]
SelfSearch (full) & \textbf{77.5} & 75.0 & \textbf{76.2} & \textbf{86.7} & \textbf{86.7} & \textbf{86.7} \\
\bottomrule
\end{tabular}
\end{table}

\subsection{How Improvements Develop}
\label{sec:results-changes}
\label{sec:lineage-interaction}
Table~\ref{tab:generation-changes} summarizes the changes made by the GPT
agents across ten generations. Episode records allow each lineage to build
on its own changes and inspect those made by the other. We observe both
direct reuse of tools and further refinement as subsequent episodes expose
limitations.

\begin{table}[t]
\caption{Changes introduced across ten generations of SelfSearch with GPT.
Repeated entries may reflect changes incorporated from the other lineage.
Bold highlights changes discussed in the analysis.}
\label{tab:generation-changes}
\centering
\small
\setlength{\tabcolsep}{5pt}
\renewcommand{\arraystretch}{1.12}
\begin{tabular}{@{}>{\raggedleft\arraybackslash}p{2em} *{2}{>{\raggedright\arraybackslash}p{\dimexpr(\linewidth-2em-4\tabcolsep)/2\relax}}@{}}
\toprule
Gen. & Capability lineage & Adaptive lineage \\
\midrule
\csname @@input\endcsname tables/generation_changes.tex
\bottomrule
\end{tabular}
\end{table}

\paragraph{GPT inspection tools.}
In generation 6, the capability lineage adds text search with output limits,
while the adaptive lineage introduces a trajectory reader. In generation 7,
each incorporates the other's tool from its episode record. The adaptive
agent uses its trajectory reader to inspect the capability lineage's record
before copying the search implementation and tests. Using the reader reveals
that combining several short excerpts can still produce an oversized
response. Generation 8 adds overall output limits. Later generations let
agents filter the records to find relevant events and see each tool result
alongside the tool call and arguments that produced it.

\paragraph{DeepSeek search-output repair.}
In generation 5, the adaptive lineage limits search output by keeping the
beginning and end of long lines, which can hide matching text in the middle.
In generation 6, the capability lineage uses that episode record to revise
the tool so that excerpts center on the match. In generation 7, the adaptive
agent reproduces the failure, incorporates the capability lineage's repair,
and verifies that matching text remains visible. Other changes preserve
information during file operations, including tabs and line endings.
Table~\ref{tab:generation-changes-deepseek} in Appendix~\ref{app:trajectories}
provides the generation-by-generation changes.

\subsection{Downstream Use of Evolved Tools}
\label{sec:analysis-transfer}
We examine whether tools developed during self-improvement are reused on
downstream tasks. We inspect both final agents on 269 tasks spanning
SWE-bench Verified, SWE-bench Multilingual, and Terminal-Bench 2.1. Averaged
across the two lineages, GPT and DeepSeek agents use their text-search tools
on 8.6\% and 23.6\% of tasks, respectively, and line-range file viewing on
40.3\% and 33.8\%. These tools are used across downstream benchmarks, while GPT's trajectory
reader is not invoked during downstream evaluation. Appendix~\ref{app:trajectories}
reports usage by tool, lineage, and benchmark (Table~\ref{tab:tool-adoption}).

\begin{figure}[t]
\centering
\setlength{\parskip}{0pt}
\setlength{\abovecaptionskip}{4pt}
\includegraphics[width=\linewidth]{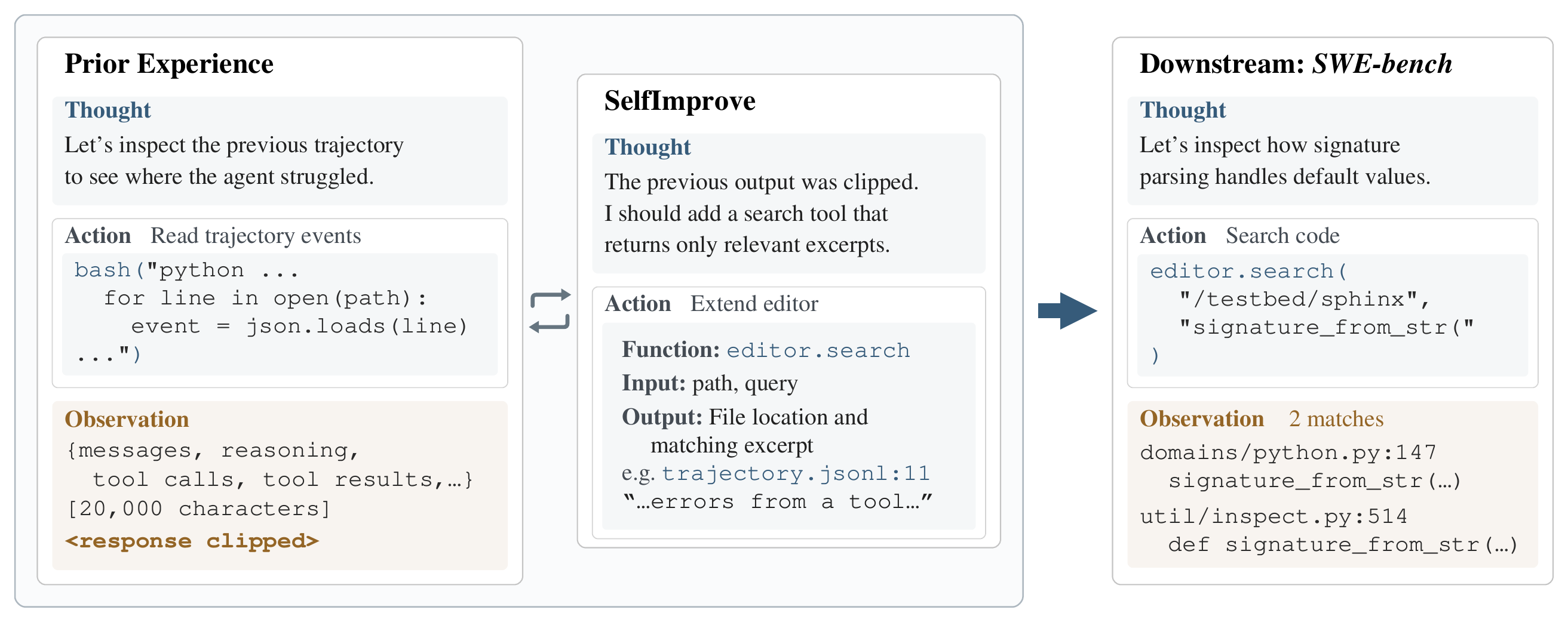}
\caption{A clipped episode record motivates a text-search tool, which is later
reused to inspect downstream task code.}
\label{fig:episode-transfer}
\end{figure}

Figure~\ref{fig:episode-transfer} illustrates this reuse in a Sphinx task.
After repairing a function, the GPT capability agent uses the text-search tool
to locate related tests and code that calls the function. It inspects the
calling code and runs the related tests, using a tool developed for
episode-record inspection to support downstream verification.

\section{Related Work}
\label{sec:related-work}

\paragraph{Evaluation-guided agent search.}
Prior work searches over agent implementations and workflows by proposing
changes, evaluating candidates, and using the resulting feedback to guide
further search~\citep{hu2024adas,zhang2025aflow,zhang2026darwin}. Archives retain
earlier designs that can serve as starting points for later revisions.
Selection can also consider an agent's potential to produce useful descendants,
rather than only its current task performance~\citep{wang2026huxleygodel}. Alongside
candidate selection, experience sharing helps agents draw on discoveries made
elsewhere in the search. Records from multiple agents, tasks, or lineages can
inform new modifications and combine complementary
improvements~\citep{weng2026gea,liu2026mgm}. SelfSearch also carries experience
across agents and generations, but obtains that experience from the
self-improvement process itself. Previous episode records guide revisions
without downstream benchmark evaluation during search.

\paragraph{Self-modifying agents.}
An agent's code determines both how it solves tasks and how it modifies other
agents. When the same agent performs both activities, revisions to its tools
and instructions can also change how it makes subsequent
improvements~\citep{robeyns2025a}. More explicitly, the improvement procedure
itself can be revised by applying it to its own implementation or allowing a
meta-agent to modify itself~\citep{zelikman2024selftaught,zhang2026hyperagents}.
These approaches also differ in when modifications take effect. Some evolve
agents across generations, while others modify the scaffold during an
individual task~\citep{xia2025liveswe}. SelfSearch uses a single agent for task
solving and self-improvement, with the entire repository editable. Revised
implementations and episode records carry forward together, so later agents
can build on both earlier changes and the experience of making them.
The operations developed by SelfSearch agents, such as line-range viewing,
exact-text replacement, and bounded search output, resemble agent--computer
interface designs shown to be effective for coding agents \citep{yang2024sweagent}.
Appendix~\ref{app:method-comparison} compares how prior systems organize these
roles and which parts can evolve. 

\paragraph{Experience from self-improvement.}
An improvement attempt produces both a revised agent and a record of the
process that produced it. Prior work uses histories of candidate changes and
evaluation outcomes to guide later revisions, retaining lessons across
generations or revising the search procedure
itself~\citep{zhang2026hyperagents,lee2026metaharness,qu2026bilevel}.
SelfSearch uses the editing process as experience: its episode records capture
the reasoning, tool calls, and intermediate outcomes involved in modifying
the agent. Failed editing actions, incomplete observations, and difficulties
inspecting earlier records can therefore motivate changes to the agent's own
tools and procedures. This allows subsequent revisions to draw on experience
gained during self-improvement, without downstream task evaluation during
search.

\section{Discussion}
\label{sec:discussion}

Downstream evaluation provides useful feedback, but repeatedly evaluating
candidates increases search cost as task coverage or repetitions grow.
Our comparison (Section~\ref{sec:evaluation-guided-comparison}) shows that
SelfSearch can produce competitive agents using self-improvement experience
instead. This separates candidate generation from downstream assessment,
while leaving evaluation and selection as separate costs.

Self-improvement can reveal limitations in the agent's own tools and
procedures. Episode records capture difficulties encountered while inspecting
code, making edits, and checking their effects, giving subsequent agents
concrete problems to investigate. Our analysis shows that addressing these
difficulties can produce tools reused on downstream tasks, as well as tools
such as the trajectory reader that support further self-improvement. Our results provide evidence that self-improvement experience can guide useful agent revisions without downstream evaluation. Further work should examine how consistently these gains arise across independent searches.

Self-improvement experience and downstream evaluation can play complementary
roles. Within an evaluation-guided search, SelfSearch could extend a single
modification step into several generations of revisions informed by episode
records. The resulting candidate would then be evaluated to determine whether
it should be retained. Future work could test whether combining SelfSearch
with evaluation-guided selection produces better agents under the same total
search budget.

\section{Conclusion}
\label{sec:conclusion}

We introduced SelfSearch, a reward-free agent search procedure that uses
records of self-improvement episodes to guide subsequent revisions. Without
downstream evaluation during search, the final agent populations improve mean
task success over their initial agents across three benchmarks, with gains
transferring across execution models. These findings suggest that the
process of editing an agent can itself provide experience for further agent
improvement.

\subsection*{AI Use Statement}

In this work, we used generative AI tools to help develop conceptual
frameworks, implement methods, refine research hypotheses, provide feedback
on experimental design, support analysis of agent trajectories, and interpret
results. We did not use generative AI tools to generate synthetic datasets,
and mathematical claims, proofs, translation, and dataset cleaning are not
applicable to this work. We also used generative AI tools to identify
related literature, create and edit code and figures, draft and edit portions
of the manuscript, and format references. We reviewed all AI-assisted work:
the authors tested AI-assisted code, checked cited works against the original
papers, and verified reported numbers and figures against experiment logs and
source data. Separately, LLM-based agents are the subject of this study, as
described in Section~\ref{sec:selfsearch}. We take responsibility for the
final content of this work, including text, claims, and artifacts produced
with the aid of generative AI.

\subsection*{Reproducibility Statement}

Section~\ref{sec:selfsearch} describes the SelfSearch procedure.
Appendices~\ref{app:harness-kernel}--\ref{app:experimental-details} provide
the agent implementation details, search prompts, model configurations,
and evaluation task selection. Our execution framework records agent
implementations and trajectories while keeping inference settings and
resource limits outside the editable agent repository.

\bibliography{main}
\bibliographystyle{iclr2027_conference}

\appendix

\section{Initial Agent and Execution Environment}
\label{app:harness-kernel}

\paragraph{Initial agent and relation to DGM.}
Our initial agent adapts DGM's coding-agent design~\citep{zhang2026darwin},
retaining shell execution and a file editor with view, create, and whole-file
replacement operations. We add an editable policy library containing
instructions for general tasks, coding, and self-improvement. At the start of
an episode, the agent receives a policy catalog and loads the instructions
and associated helper tools relevant to its task. The same agent implementation
handles downstream tasks and self-modification.

\paragraph{Editable repository and fixed runtime.}
SelfSearch can modify the entire agent repository, including its instructions,
tools, policy library, and task-level orchestration. The agent invokes the
runtime through \texttt{respond(message)}, which executes the model--tool
interaction loop using the agent's instructions and tool definitions. The
runtime remains outside the editable repository and fixes the model, reasoning
effort, generation settings, and execution limits. Model-call and tool-step
limits apply across the entire run, including multiple calls to
\texttt{respond(message)}.

\paragraph{Runtime-managed episode records.}
The runtime records the agent's trajectory, including reasoning, tool actions,
and outcomes, outside the editable repository. The search workflow stores this
trajectory alongside the agent's source code before and after modification.
Previous episode records are mounted read-only during subsequent episodes.

\paragraph{Portability.}
We evaluate the same frozen agent implementation across benchmarks without
modifying its source code. Benchmark adapters provide task instructions and
access to the task environment, then pass the agent's outputs to the
corresponding evaluator. Environment setup and grading remain outside the
agent repository.

\section{Search Prompts}
\label{app:selfsearch}
Each search prompt combines the shared template with a model-specific goal
and a capability or adaptive search direction. The blocks below fill the
corresponding placeholders, with goal and search-direction headings omitted.
The agent's system instructions and policy catalog are supplied separately.
The goal and design-guidance texts were refined over pilot searches by inspecting improver trajectories and code changes, without evaluating pilot agents on downstream benchmarks, and were fixed before the searches reported here.
% Generated from frozen plans by scripts/build_prompt_appendix.py.
\subsubsection*{SelfSearch prompt template}
\begin{promptbox}
\# Objective\par
\smallskip
\{goal\}\par
\smallskip
\{design\_guidance\}\par
\smallskip
\{search\_direction\}\par
\smallskip
\hangindent=1.2em\hangafter=1 - A change should increase useful capability or reliability, not merely keep the package runnable.\par
\smallskip
\hangindent=1.2em\hangafter=1 - When evidence is available, use observed behavior and outcomes to decide what to keep, revise, or remove.\par
\smallskip
\hangindent=1.2em\hangafter=1 - Keep behavior broadly useful across tasks. Do not turn one execution's circumstances into universal requirements.\par
\smallskip
\hangindent=1.2em\hangafter=1 - Let the agent interpret task intent and choose relevant policies and tools from context.\par
\smallskip
\hangindent=1.2em\hangafter=1 - Use additional `self.respond` passes judiciously; each adds substantial recurring cost.\par
\smallskip
\# Environment\par
\smallskip
\hangindent=1.2em\hangafter=1 - The agent package must export `agent` from `\_\_init\_\_.py`. The exported object must be an instance of `Agent`.\par
\smallskip
\hangindent=1.2em\hangafter=1 - `agent.run(task: str)` receives the current task and must return a string. Each evaluation imports the returned package in a fresh isolated execution.\par
\smallskip
\hangindent=1.2em\hangafter=1 - The agent's `system\_prompt` is supplied as the system message. `self.respond(message)` advances the active model-and-tool conversation.\par
\smallskip
\hangindent=1.2em\hangafter=1 - After `agent.run()` returns successfully, the host exports the final agent in `/workspace` and checks its contract.\par
\smallskip
\hangindent=1.2em\hangafter=1 - `@tool` defines a tool's schema through its name, type annotations, and docstring. Tools registered in `agent.tools` are directly model-visible whenever that agent is executed.\par
\smallskip
`/app/agent` is the read-only agent currently executing. `/workspace` begins as the same source and is editable. Workspace edits affect the returned successor, not the current execution.\par
\smallskip
\# Procedure\par
\smallskip
`/workspace` contains the current agent. `/evidence/0` and `/evidence/1` contain two previous editing executions.\par
\smallskip
Start at `/evidence/index.json`, which links factual summaries, final accounts, edits, trajectories, and source snapshots. Determine the most consequential limitation or opportunity supported by the evidence. Develop one coherent revision that addresses it. The revision may span multiple components when the underlying change requires it.\par
\smallskip
Edit and verify the agent.\par
\smallskip
Return a concise account and leave the resulting agent in `/workspace`.\par
\end{promptbox}
\subsubsection*{Capability direction}
\begin{promptbox}
\# Search direction: capability expansion\par
\smallskip
Identify a consequential task-solving limitation supported by the available evidence, then develop a reusable capability that addresses it. Prefer a working mechanism over additional instructions when instructions alone cannot supply the missing operation, and avoid specializing the agent to self-improvement episodes.\par
\end{promptbox}
\subsubsection*{Adaptive direction}
\begin{promptbox}
\# Search direction: adaptive reasoning\par
\smallskip
Improve how the agent changes its approach when evidence contradicts its current plan or supports a better one. Preserve flexibility across tasks rather than routing behavior through keywords or a fixed workflow.\par
\end{promptbox}
\subsubsection*{GPT goal}
\begin{promptbox}
\# Goal\par
\smallskip
Produce an executable agent with greater ability to improve agents, including itself.\par
\end{promptbox}
\subsubsection*{DeepSeek goal}
\begin{promptbox}
\# Goal\par
\smallskip
Produce an executable agent with greater expected task-solving capability, including the ability to improve agents.\par
\end{promptbox}
\subsubsection*{DeepSeek design guidance}
\begin{promptbox}
\# Design guidance\par
\smallskip
Prioritize changes that improve the agent's behavior during ordinary task execution.\par
\smallskip
When adding or changing a model-visible tool, verify the exact call shape the agent is expected to produce. Prefer interfaces that the model can use reliably; a valid implementation or schema alone is not sufficient.\par
\smallskip
Keep planning, review, and persistent state proportional to the task. Do not make additional bookkeeping or repeated passes mandatory unless observed behavior shows that their benefit justifies their cost.\par
\smallskip
Preserve broadly useful existing behavior and avoid reproducing host search, evaluation, or orchestration inside the returned agent.\par
\end{promptbox}

\section{Experimental Details}
\label{app:experimental-details}

\subsection{Model Configurations}
Search uses \texttt{gpt-5.6-sol} for the GPT configuration and
\texttt{deepseek-v4-pro-0813} for the DeepSeek configuration, both with medium
reasoning effort and temperature 1. During search, per-call output limits are
8,192 and 16,384 tokens, respectively. Each self-improvement episode allows
up to 513 model calls, 512 tool steps, and four hours of execution in a
container without network access.

Downstream evaluation uses \texttt{gpt-5.6-luna} and
\texttt{deepseek-v4-flash-0731}, both with medium reasoning effort,
temperature 1, and a 16,384-token per-call output limit. The comparison
with other harnesses uses the settings described in
Appendix~\ref{app:published-harness-comparison}.

\subsection{Evaluation Sets and Protocol}
We construct the 120-task SWE-bench Verified evaluation set by combining the
60 tasks in DGM's small and medium subsets~\citep{zhang2026darwin} with 60 additional tasks sampled from
its large subset. The additional task IDs are listed in
Table~\ref{tab:additional-verified-tasks}.
For SWE-bench Multilingual, we sample 60 tasks using seed 42, with
C (6), C++ (2), Go (8), Java (9), JavaScript (9), PHP (9), Ruby (8),
and Rust (9). Table~\ref{tab:multilingual-tasks} lists the sampled task IDs.
We evaluate on all 89 tasks in Terminal-Bench 2.1.

\begin{table}[t]
\caption{The 60 additional SWE-bench Verified task IDs sampled from DGM's large subset.}
\label{tab:additional-verified-tasks}
\centering
\footnotesize
\setlength{\tabcolsep}{8pt}
\begin{tabular}{@{}lll@{}}
\toprule
\multicolumn{3}{c}{Task IDs} \\
\midrule
\csname @@input\endcsname tables/additional_verified_tasks.tex
\bottomrule
\end{tabular}
\end{table}

\begin{table}[t]
\caption{The 60 sampled SWE-bench Multilingual task IDs.}
\label{tab:multilingual-tasks}
\centering
\footnotesize
\setlength{\tabcolsep}{8pt}
\begin{tabular}{@{}ll@{}}
\toprule
\multicolumn{2}{c}{Task IDs} \\
\midrule
\csname @@input\endcsname tables/multilingual_tasks.tex
\bottomrule
\end{tabular}
\end{table}

\subsection{Evaluation-Guided Search Comparison}
\label{app:evaluation-guided}
Both baselines perform ten revisions, with the selected parent editing a copy
of itself. Linear search always uses the latest candidate. Archive search
retains $B_0$ and all completed candidates. After the shared first revision,
it samples a parent uniformly on even-numbered revisions and chooses the
highest-scoring parent on odd-numbered revisions, breaking ties toward the
earliest candidate.

The editor receives its parent's execution records on DGM's ten-task small
subset, including task statements, trajectories, submitted patches, outcomes,
and official tests and test outputs. Official tests are available only after
task execution. Both baselines ask for reusable capability improvements.
Local checks are allowed during editing, and the returned candidate is then
evaluated on the development set.

Final selection uses the highest development score among the ten revisions,
breaking ties toward the latest. The selected linear and archive revisions
are 8 and 10 for GPT, and 6 and 9 for DeepSeek. SelfSearch instead reports both
generation-10 lineages without development-score selection, so the comparison
is not budget-matched.

We exclude the ten development tasks from every method's SWE-bench Verified evaluation,
leaving 110 tasks, and also evaluate on SWE-bench Multilingual. The benchmark grader
determines success, including when execution terminates after producing a
valid patch.

\subsection{Cost Metrics and Inference Pricing}
\label{app:pricing}
Let $c_i(B)$ denote the total model execution cost of agent $B$ on task $i$,
and let $s_i(B)\in\{0,1\}$ indicate whether it solves the task. For an
evaluation set $\mathcal T$, average execution cost per task is
\begin{equation}
\bar c(B)=\frac{1}{|\mathcal T|}\sum_{i\in\mathcal T}c_i(B).
\end{equation}
This average includes successful and unsuccessful tasks. The percentage cost
reduction relative to the initial agent is
\begin{equation}
R(B)=100\left(1-\frac{\bar c(B)}{\bar c(B_0)}\right).
\end{equation}
To compare costs on tasks solved by both agents, define
$\mathcal S_B=\{i\in\mathcal T:s_i(B)=s_i(B_0)=1\}$. We compute
\begin{equation}
R_{\mathrm{shared}}(B)=100\left(
1-\frac{\sum_{i\in\mathcal S_B}c_i(B)}
{\sum_{i\in\mathcal S_B}c_i(B_0)}\right).
\end{equation}
Both agents are compared on the same tasks within each pair. Positive values
indicate cost reductions, and negative values indicate increases. These
execution-cost metrics exclude search cost.

Per million tokens, input, cached-input, and output prices are \$0.20,
\$0.02, and \$1.20 for GPT-5.6 Luna, and \$0.15, \$0.003, and \$0.60 for
DeepSeek V4 Flash. Input prices apply to uncached tokens.

\clearpage
\section{Additional Results}
\label{app:complete-results}

\subsection{Search Cost}
\label{app:search-cost}
Table~\ref{tab:search-cost} reports model calls, token usage, and inference
costs for the initialization and ten-generation segments of SelfSearch.

\begin{table}[h]
\caption{Search resources. Input tokens include cached input. M and k denote millions and thousands of tokens. Costs are in USD and exclude downstream evaluation.}
\label{tab:search-cost}
\centering\tablestyle
\begin{tabular}{@{}lrrrr@{}}
\toprule
Model & Calls & Input (M) & Output (k) & Total cost (\$) \\
\midrule
GPT-5.6 Sol & 281 & 5.48 & 101.4 & 6.52 \\
DeepSeek V4 Pro & 832 & 36.89 & 578.7 & 4.03 \\
\bottomrule
\end{tabular}
\end{table}

\subsection{Downstream Efficiency}
\begin{figure}[htbp]
\centering
\includegraphics[width=\linewidth]{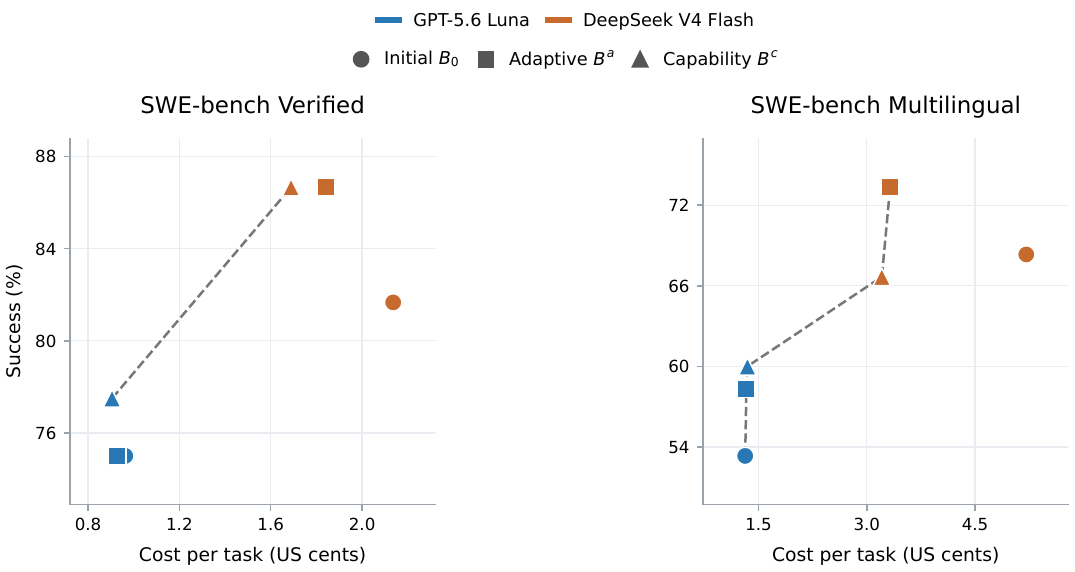}
\caption{Accuracy--cost tradeoffs on SWE-bench Verified and SWE-bench
Multilingual. Costs and markers follow Figure~\ref{fig:terminal-cost-frontier}.
Dashed lines connect non-dominated evaluated agents in each subplot.
Axis ranges differ between benchmarks.}
\label{fig:swe-cost-frontier}
\end{figure}

Figure~\ref{fig:swe-cost-frontier} shows the accuracy--cost tradeoffs on
SWE-bench Verified and SWE-bench Multilingual. Table~\ref{tab:efficiency}
reports average model calls and token usage across all tasks, including unsuccessful
attempts. Table~\ref{tab:common-success} compares costs on tasks solved by
both the initial and evolved agents, allowing us to examine execution
efficiency on their shared successes.

\begin{table}[H]
\caption{Mean execution resources per task, including unsuccessful tasks.
Tokens are in thousands, and input includes cached tokens.}
\label{tab:efficiency}
\label{tab:efficiency-gpt}
\label{tab:efficiency-deepseek}
\centering
\tablestyle
\resizebox{\linewidth}{!}{\begin{tabular}{@{}llrrrrrr@{}}
\toprule
 & & \multicolumn{3}{c}{GPT-5.6 Luna} & \multicolumn{3}{c}{DeepSeek V4 Flash} \\
\cmidrule(lr){3-5}\cmidrule(l){6-8}
Benchmark & Agent & Calls & Input (k) & Output (k) & Calls & Input (k) & Output (k) \\
\midrule
\csname @@input\endcsname tables/efficiency_combined_rows.tex
\end{tabular}}
\end{table}

\begin{table}[H]
\caption{Execution cost reductions relative to $B_0$ on tasks solved by both
agents. Negative values indicate increased cost.}
\label{tab:common-success}
\label{tab:common-success-gpt}
\label{tab:common-success-deepseek}
\centering\tablestyle
\resizebox{\linewidth}{!}{\begin{tabular}{@{}llrrrr@{}}
\toprule
 & & \multicolumn{2}{c}{Capability ($B^{c}$)} & \multicolumn{2}{c}{Adaptive ($B^{a}$)} \\
\cmidrule(lr){3-4}\cmidrule(l){5-6}
Model & Benchmark & Shared tasks & Reduction (\%) & Shared tasks & Reduction (\%) \\
\midrule
\csname @@input\endcsname tables/common_success_combined_rows.tex
\end{tabular}}
\end{table}

\subsection{Comparison with Other Harnesses}
\label{app:published-harness-comparison}
We evaluate the capability agent $B^{c}$ found by SelfSearch using DeepSeek
V4 Pro on all 89 Terminal-Bench 2.1 tasks. Execution uses DeepSeek V4 Flash,
matching the model and task set in the nine-harness comparison of
\citet{apachemaka2026ninearm}. We use
\texttt{xhigh} reasoning, each task's predefined time limit, and limits of 100,000 tool
steps and 100,001 model calls. SelfSearch and Codex each solve 73 tasks,
including 66 solved by both and seven solved only by each harness.

\section{Self-Improvement Trajectories}
\label{app:trajectories}

\subsection{Changes across Generations}

\paragraph{GPT trajectory-reader sequence.}
The adaptive lineage introduces \texttt{inspect\_trajectory} in generation 6
to read previous episode records. The capability lineage incorporates it in
generation 7. In generation 8, it finds that limiting
individual excerpts still allows overly long responses, so it adds an overall
response limit and a continuation position for retrieving the remaining content.
Generation 9 uses this interface to inspect both previous trajectories and
adds filtering. Generation 10 links tool results to their originating calls
and arguments, including across response pages. Each generation uses and
refines the inspection tool inherited from its predecessor.

\begin{table}[htbp]
\caption{Changes introduced across ten generations of SelfSearch with DeepSeek.
Repeated entries may reflect changes incorporated from the other lineage.
Bold highlights changes discussed in the analysis.}
\label{tab:generation-changes-deepseek}
\centering
\small
\setlength{\tabcolsep}{5pt}
\renewcommand{\arraystretch}{1.12}
\begin{tabular}{@{}>{\raggedleft\arraybackslash}p{2em} *{2}{>{\raggedright\arraybackslash}p{\dimexpr(\linewidth-2em-4\tabcolsep)/2\relax}}@{}}
\toprule
Gen. & Capability lineage & Adaptive lineage \\
\midrule
\csname @@input\endcsname tables/generation_changes_deepseek.tex
\bottomrule
\end{tabular}
\end{table}

\paragraph{DeepSeek observation and editing sequence.}
Table~\ref{tab:generation-changes-deepseek} lists the changes across all ten
generations.
DeepSeek's capability lineage develops tools for inspecting long outputs,
then repairs problems exposed by their use. Generation 3 adds head-and-tail
excerpts for shell output, generation 4 fixes a temporary-file collision,
and generations 5--7 extend output handling to file views, search results,
and directory listings. The adaptive lineage uses the other lineage's
records to preserve matching text when shortening search results. Later
capability generations repair inconsistencies between file viewing and
editing by preserving tabs and line endings.

\subsection{Downstream Tool Use}
Table~\ref{tab:tool-adoption} reports the percentage of downstream tasks on
which each final agent invokes an introduced tool or operation. Each task is
counted once per operation, regardless of the number or success of its calls.
The ``All'' column pools the three benchmarks, weighting each task equally.
Related refinements are grouped under the same operation. Shell commands such
as \texttt{grep} are not counted as uses of the introduced search tool.

\begin{table}[H]
\caption{Downstream use of introduced operations by the final capability and
adaptive agents. Entries give the percentage of tasks with at least one invocation.
Character-range viewing and the trajectory reader were not introduced in the
DeepSeek agents and are omitted for those agents.}
\label{tab:tool-adoption}
\centering
\tablestyle
\setlength{\tabcolsep}{3pt}
\begin{tabular}{@{}llrrrr@{}}
\toprule
Agent & Operation & \shortstack{SWE-bench\\Verified} & \shortstack{SWE-bench\\Multilingual} & \shortstack{Terminal-\\Bench 2.1} & All \\
\midrule
\csname @@input\endcsname tables/tool_adoption_rows.tex
\bottomrule
\end{tabular}
\end{table}

\section{Comparison of Agent Search Designs}
\label{app:method-comparison}

Table~\ref{tab:method-comparison} compares the evolution of task and
self-improvement roles, their organization, and the use of downstream rewards.

\begin{table}[ht]
\caption{Comparison of agent search designs. A checkmark denotes the property
as defined below, and a dash denotes its absence.}
\label{tab:method-comparison}
\centering
\tablestyle
\begin{tabular}{@{}lcccc@{}}
\toprule
\textbf{Method}
& \shortstack{\textbf{Evolving}\\\textbf{task agent}}
& \shortstack{\textbf{Evolving}\\\textbf{meta-agent}}
& \shortstack{\textbf{Unified meta}\\\textbf{and task agent}}
& \shortstack{\textbf{Reward-free}\\\textbf{search}} \\
\midrule
DGM~\citep{zhang2026darwin} & $\checkmark$ & --- & --- & --- \\
HGM~\citep{wang2026huxleygodel} & $\checkmark$ & --- & --- & --- \\
Hyperagents~\citep{zhang2026hyperagents} & $\checkmark$ & $\checkmark$ & --- & --- \\
SICA~\citep{robeyns2025a} & $\checkmark$ & $\checkmark$ & $\checkmark$ & --- \\
SelfSearch (ours) & $\checkmark$ & $\checkmark$ & $\checkmark$ & $\checkmark$ \\
\bottomrule
\end{tabular}
\end{table}

An evolving task agent changes across search generations in how it solves
downstream tasks. An evolving meta-agent changes how it proposes and implements
agent modifications. The roles are unified when the same agent solves tasks
and modifies itself. Reward-free search does not use downstream evaluation
rewards to guide revisions or selection.

DGM and HGM use a separate diagnosis procedure to propose
modifications~\citep{zhang2026darwin,wang2026huxleygodel}, while Hyperagents defines
distinct task and meta-agents within an editable
program~\citep{zhang2026hyperagents}.

\end{document}